\documentclass[letterpaper, 10 pt, conference]{ieeeconf} 

\IEEEoverridecommandlockouts                              
\usepackage{amsmath}
\usepackage{subcaption}
\usepackage{booktabs} 
\usepackage{caption}  
\usepackage{tabularx}
\usepackage{multirow}
\usepackage{graphicx}
\usepackage{array}
\usepackage[export]{adjustbox}
\usepackage{xcolor}
\usepackage{hyperref}
\usepackage[normalem]{ulem}

\newcommand{\rev}[1]{{\color{black}#1}}
\newcommand{\sys}{\rev{RoboHarness}}
\newcommand{\titleemoji}{%
  \raisebox{-0.18ex}{\includegraphics[height=1.25em,trim=220 80 150 80,clip]{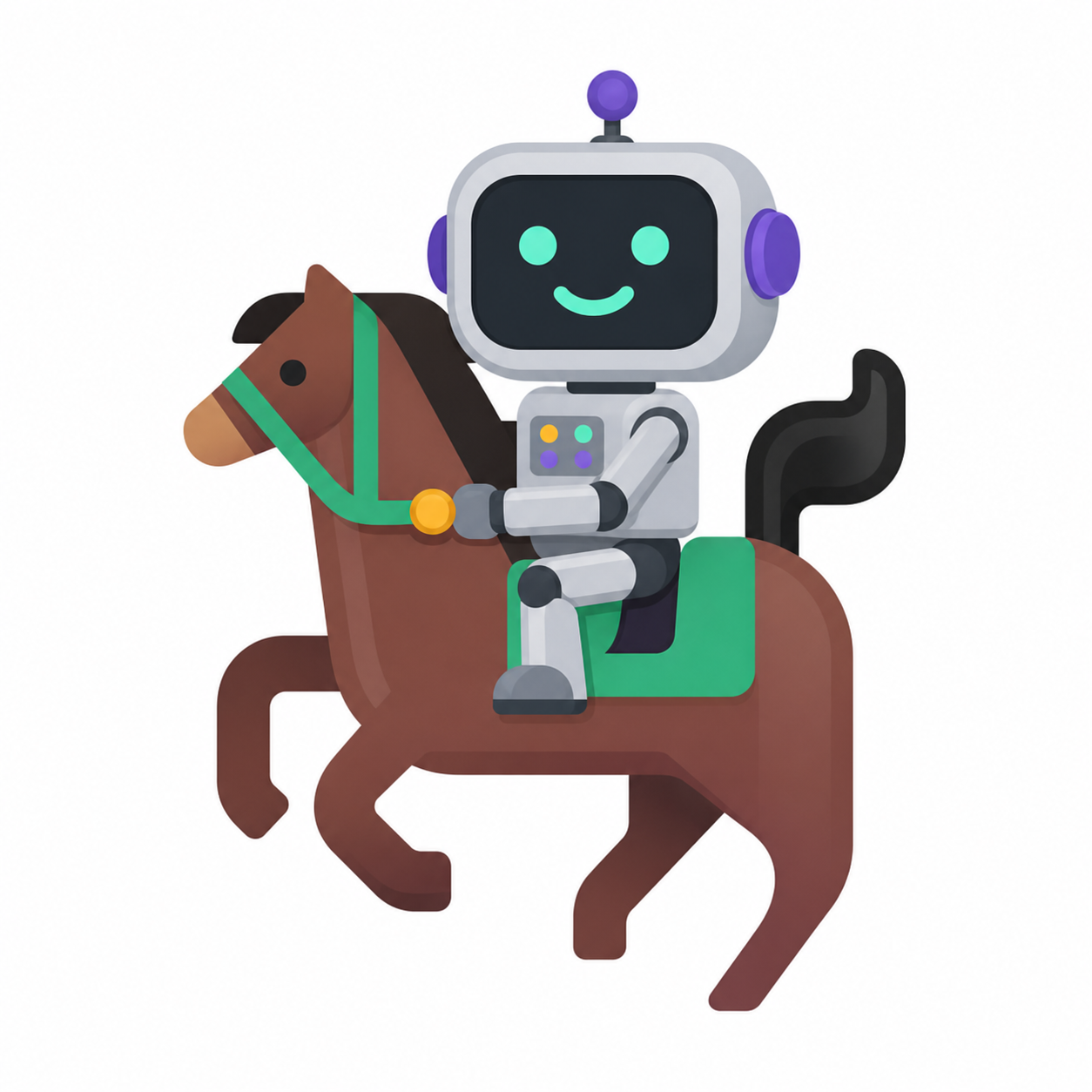}}%
}
\title{\LARGE \bf
\rev{RoboHarness~\titleemoji:\@ A Memory-Augmented Policy Harness for Vision-Language-Action Model Robustness via In-Context Adaptation}
}

\author{
    \authorblockN{Zhuoran Li$^{1,*}$, Zhiyang Li$^{1,*}$, Kaijun Zhou$^{1}$,  Jinyu Gu$^{1,2,\dagger}$} \\
}

\begin{document}

\twocolumn[{
    \renewcommand\twocolumn[1][]{#1}
    \maketitle 
    \thispagestyle{empty}

    \vspace{-4em}
    \begin{center}
        \centering
        \includegraphics[width=0.95\textwidth]{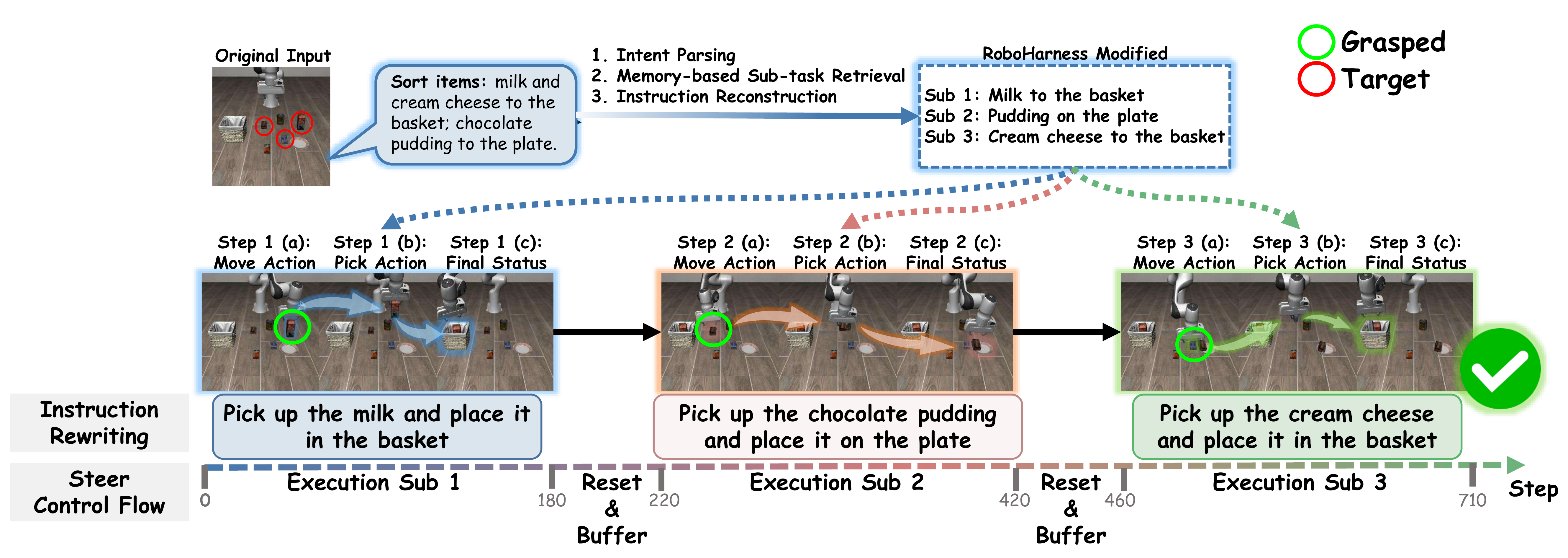}
        \captionof{figure}{\textbf{Memory-Driven Decomposition for Long-Horizon Manipulation.}  
        \sys\ transforms abstract multi-step instructions into executable sequences through intent parsing, memory retrieval, and adaptive control-flow regulation.}
        \label{fig:long_horizon}
    \end{center}
    
}]

\pagestyle{empty}

\begin{abstract}
Despite the promise of Vision-Language-Action (VLA) models as generalist robotic controllers, 
their robustness against perceptual noise and environmental variations 
in out-of-distribution (OOD) tasks remains fundamentally limited 
by the absence of long-term memory, causal failure attribution, and dynamic intervention capability.
To address this, we propose {\sys}, \rev{a memory-augmented policy harness
that upgrades frozen VLA policies for robust in-context adaptation without parameter fine-tuning.}
Specifically, {\sys} operates through an online pipeline of contrastive Dual-Memory Retrieval-Augmented Generation (RAG), 
an {\color{black}attribution-driven vision-language orchestrator implemented with a multimodal large language model}, and extensible Model Context Protocol (MCP) interventions,
while an offline Memory Consolidation module continuously distills the execution traces into reliable priors.
Experimental evaluations across three backbone models ($\pi_0$, $\pi_{0.5}$, and SmolVLA) 
on \textbf{LIBERO-PRO} and our proposed \textbf{LIBERO-{\sys}} benchmarks demonstrate 
that {\sys} achieves an average absolute success rate gain of $\mathbf{56.6\%}$. 
This includes a significant absolute improvement of $\mathbf{89.1\%}$ in long-horizon task chaining. Project page and source code are available at: \url{https://github.com/LZY-1021/RoboHarness}.
% \end{abstract}
\end{abstract}

\section{INTRODUCTION}
% --- 在这里手动插入首页脚注 ---
% 使用 fnsymbol 切换到 * 和 † 符号
\renewcommand{\thefootnote}{\fnsymbol{footnote}}
\footnotetext[1]{Equal contribution.} % 对应 *
\footnotetext[2]{Correspondence to: Jinyu Gu {\tt\small <gujinyu@sjtu.edu.cn>}.} % 对应 †

% 使用数字切换回机构脚注
\renewcommand{\thefootnote}{\arabic{footnote}}
\footnotetext[1]{School of Computer Science, Shanghai Jiao Tong University, Shanghai, China.}
\footnotetext[2]{Shanghai Syslong Information Technology Co., Ltd.}
% ---------------------------

VLA models have emerged as a leading paradigm for end-to-end robotic manipulation, unifying multimodal perception, linguistic grounding, and continuous motor control within a single learnable policy~\cite{VLA_review1, VLA_review2}.
{\color{black}In this paper, multimodal refers specifically to the joint use of visual observations or keyframes with natural-language instructions for grounding, retrieval, and intervention planning, rather than an arbitrary set of robot sensor modalities.}
In deployment, these models serve as monolithic sensorimotor controllers that directly translate visual observations and natural language instructions into low-level motor commands.

However, as robotic agents move from controlled laboratories to open-world environments, the central challenge shifts from in-distribution accuracy to robust in-context adaptation~\cite{Yuan2025FromST}.
Although fine-tuning is effective in specialized settings, general-purpose robots face interference and noise that cannot be fully anticipated during training.
Variations in object appearance, lighting, and language often induce OOD perturbations, exposing the limits of static VLA policies and motivating robustness to environmental noise and causal distractions~\cite{RT-2, hill2020humaninstructionfollowingdeepreinforcement, pmlr-v205-shah23b, liberopro,LIBERO-Plus}.

Existing approaches address OOD generalization only partially.
Scaling-centric foundation models~\cite{pi0, openvla, openx} seek generalization from large data but remain stateless controllers, discarding execution history and lacking the \textbf{long-term memory} needed for OOD adaptation.
Retrieval-augmented frameworks~\cite{ricl, memer} use past demonstrations but suffer from a ``success bias'': by imitating successes, they lack the \textbf{causal attribution} needed to diagnose failures from perceptual discrepancies.
Both paradigms treat knowledge as static, allowing early misattributions to persist across tasks.
Recent agentic frameworks~\cite{cap, voxposer, moka} add {\color{black}LLM/VLM-based planning}, but still act as zero-start reasoners with rigid tool integration, lacking the \textbf{dynamic orchestration} needed to intervene in the perception-action stream.

% Despite these advances, a common implicit assumption remains: sufficiently capable models can ``learn away'' generalization failures.
% However, in open-world settings, system failures rarely arise from isolated perceptual errors; instead, they emerge from complex couplings among task misinterpretation, perceptual uncertainty, execution errors, and environmental feedback.
% Addressing such failures requires systems to possess
These observations indicate that achieving robust OOD generalization requires a unified framework that effectively integrates:
  \emph{(i)} long-term memory for accumulating and reusing past execution experience,
  \emph{(ii)} causal attribution and failure-aware strategic adjustment,
  \emph{(iii)} dynamic orchestration of external tools to actively intervene in the perception-action stream, and
  \emph{(iv)} a self-evolving consolidation mechanism that continuously refines stored experiences to prevent knowledge stagnation.

\begin{figure}[t]
\vspace{0.5em}
     \centering
     \captionsetup[subfigure]{font=footnotesize}
     \begin{subfigure}[b]{0.15\textwidth}
         \centering
         \includegraphics[width=\textwidth]{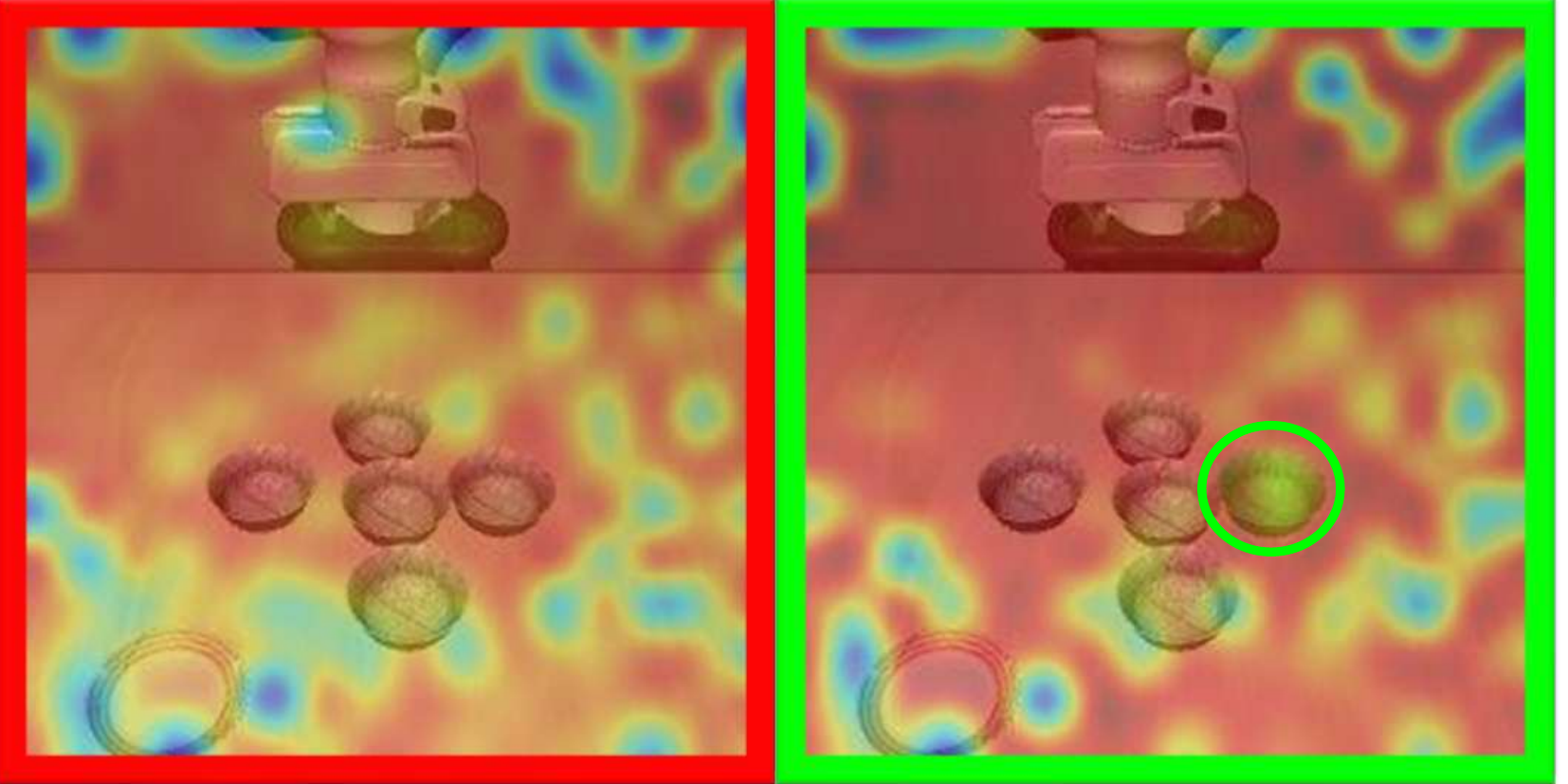}
         \caption{Visual Overlay}%
         \label{fig:sub1}
     \end{subfigure}
     \hfill
     \begin{subfigure}[b]{0.15\textwidth}
         \centering
         \includegraphics[width=\textwidth]{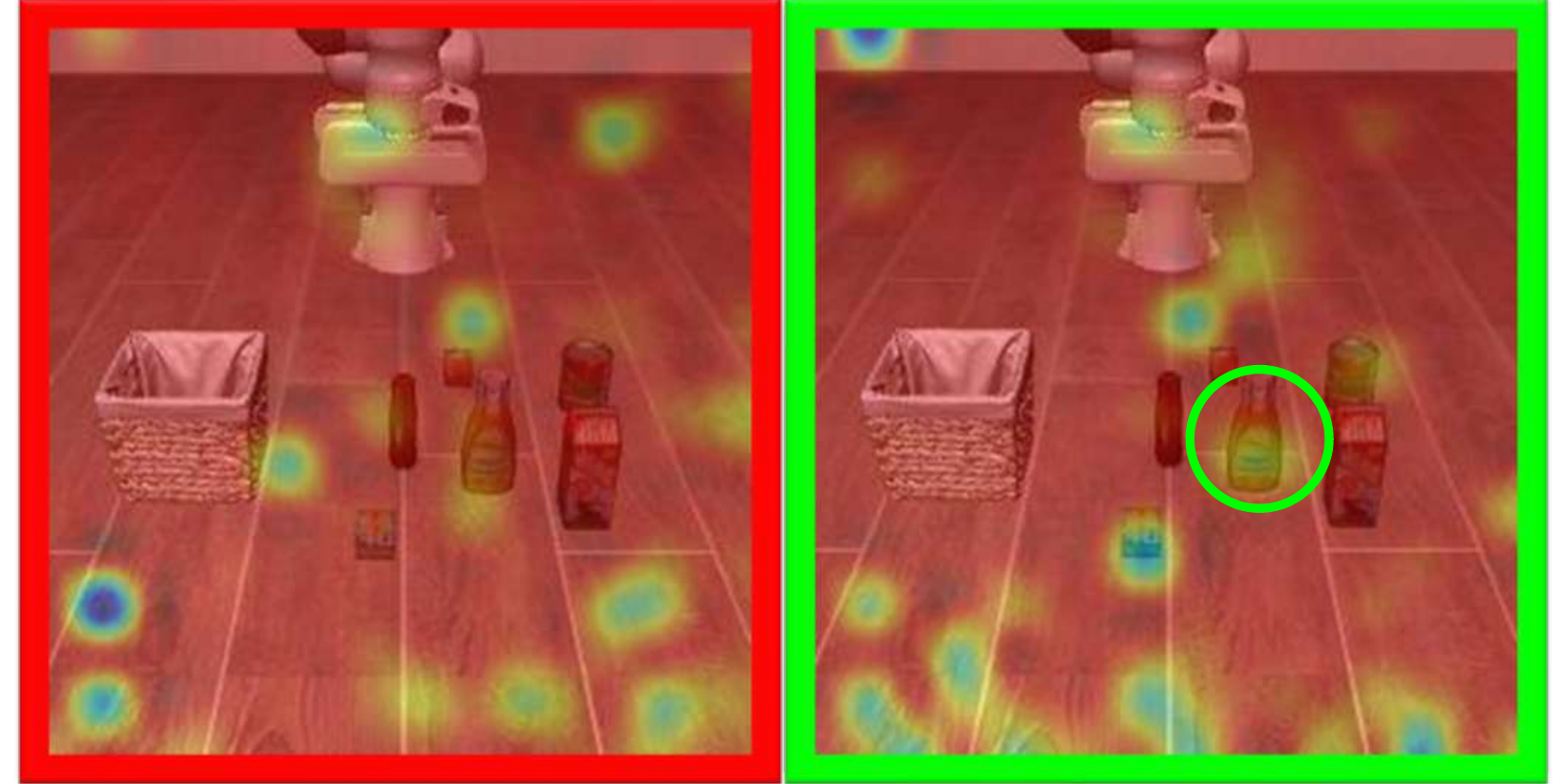}
         \caption{Prompt Simplify}%
         \label{fig:sub2}
     \end{subfigure}
     \hfill
     \begin{subfigure}[b]{0.15\textwidth}
         \centering
         \includegraphics[width=\textwidth]{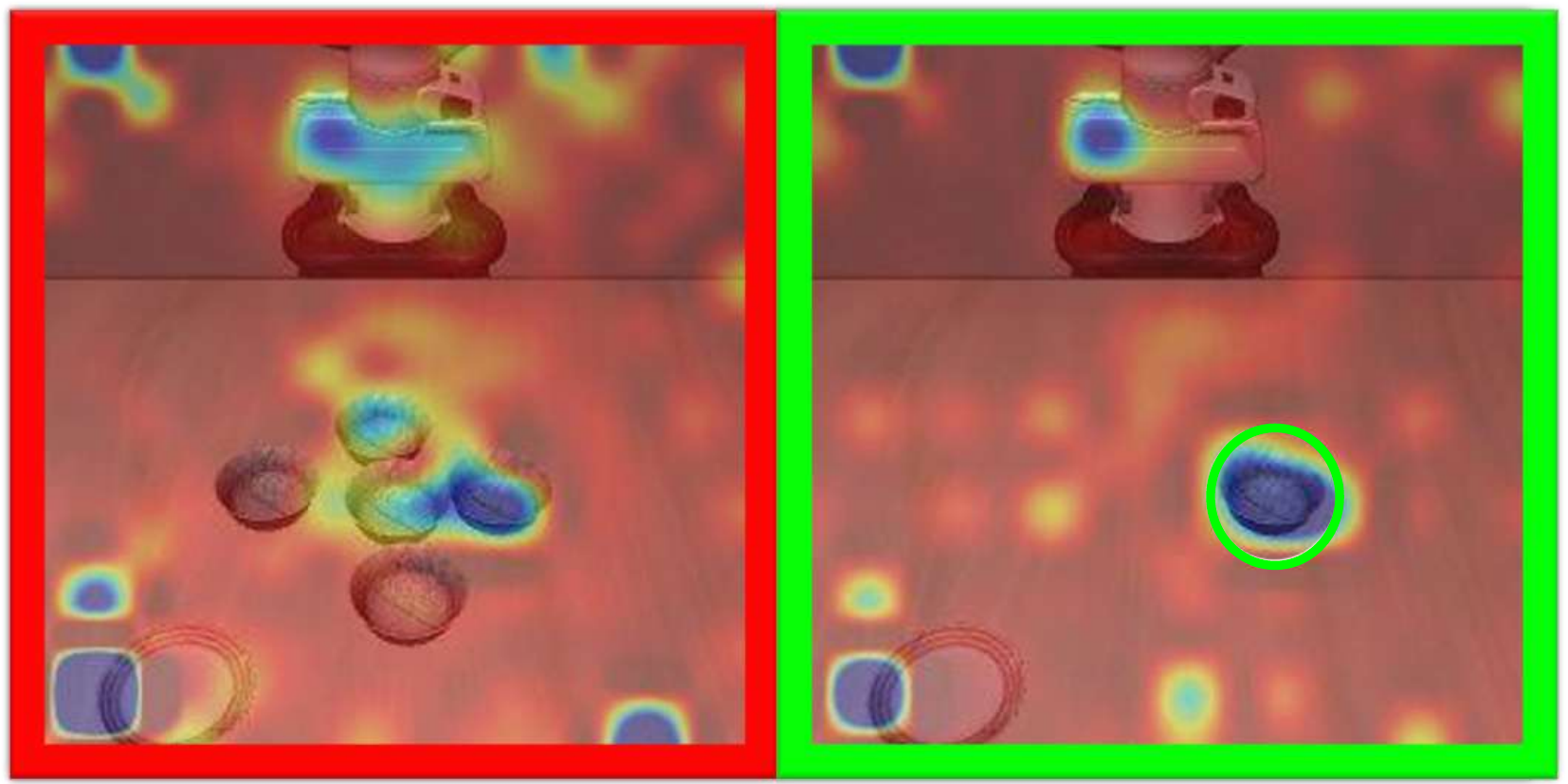}
         \caption{Distractor Remove}%
         \label{fig:sub3}
     \end{subfigure}
     \hfill
     \begin{subfigure}[b]{0.45\textwidth}
         \centering
         \includegraphics[width=\textwidth]{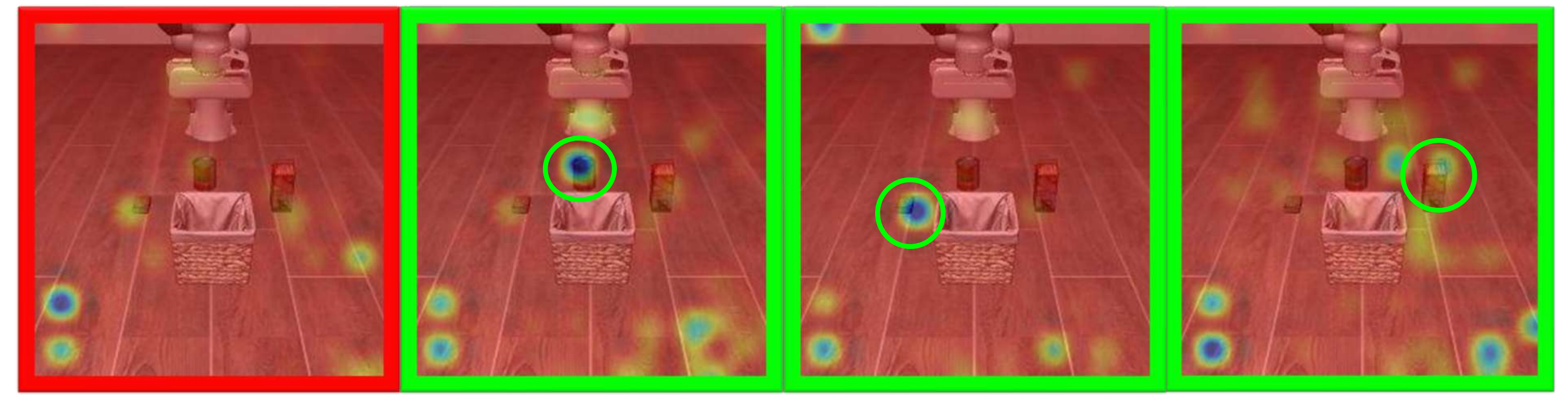}
         \caption{Long-Horizon}%
         \label{fig:sub4}
     \end{subfigure}

    \caption{\textbf{Attention maps under varied inputs.} Red/blue denote high/low attention; {\color{black}green circles mark interventions, not heatmap values.}
     Baselines (red frames) show diffuse focus, while MCP interventions (green frames) improve object localization and boundary delineation via visual cues, prompt simplification, clutter filtering, and subtask decomposition.}
     \label{fig:attention_maps}
     \vspace{-1.5em}
\end{figure}

% \rev{To address these limitations, we formulate \textbf{{\sys}} as a memory-augmented policy harness for frozen VLA models.}
% \rev{Rather than modifying policy parameters, the harness wraps the base policy with retrieval-grounded diagnosis,}
% \rev{tool-mediated perception-language intervention, and control-flow recovery, thereby regulating the observation,}
% \rev{instruction, and execution context under which the policy is queried.}
% {\sys} operationalizes the aforementioned requirements through a structurally decoupled architecture, 
% featuring three tightly integrated online components for task intervention, 
% complemented by an asynchronous offline dual-stage memory consolidation that drives continuous long-term capability improvement.

\rev{To address these limitations, we formulate \textbf{{\sys}} as a memory-augmented policy harness for frozen VLA models.}
\rev{Instead of updating parameters, it wraps the base policy with retrieval-grounded diagnosis, tool-mediated perception-language intervention, and control-flow recovery, regulating the observation, instruction, and execution context at query time.}
{\sys} implements these requirements through a decoupled architecture: three online components for task intervention and an asynchronous offline dual-stage memory consolidation module for long-term capability refinement.

\emph{(i)} \textbf{Long-Term Experience Alignment:}
As shown in Fig.~\ref{fig:attention_maps}, OOD failures often arise from perceptual ``attention drift'' 
rather than motor limits: environmental or linguistic noise disrupts the frozen VLA's focus.
To counter this, {\sys} maintains a \textbf{long-term memory bank} of successful and failed executions, 
recalling input-level adjustments that resolved similar failures.
Via MCP tool interventions~\cite{MCP}, {\sys} aligns current perceptions with past experience without parameter updates.

\emph{(ii)} \textbf{Causal Failure Attribution:}
For failures beyond retrieval, 
{\sys} uses \textbf{contrastive dual-memory retrieval} over successes and failures.
A {\color{black}vision-language orchestrator} compares them to identify perceptual ambiguity or planning deviation and guide targeted interventions.

\emph{(iii)} \textbf{Dynamic Tool Orchestration:}
The MCP converts diagnoses into \textbf{dynamically orchestrated} modular tool chains.
Based on causal diagnosis, the {\color{black}orchestrator} composes MCP sequences to rectify visual inputs or refine task context.

\emph{(iv)} \textbf{Memory Consolidation:}
After each task, an offline workflow updates memory without interrupting execution.
\textbf{Initial Diagnosis} attributes the latest trajectory and commits it to the Dual-Memory Bank; 
\textbf{Memory Consolidation} retrieves similar histories and uses cross-task differential analysis to refine stored attributions and intervention plans.
This self-correction loop prevents knowledge stagnation and improves future interventions.

Empirical evaluations across the standard \textbf{LIBERO-PRO}~\cite{liberopro} benchmark and our custom \textbf{LIBERO-{\sys}} suite demonstrate that our framework consistently enhances in-context adaptation for various VLA models without fine-tuning. 
On LIBERO-{\sys}, where base models often fail under sensory and linguistic noise, {\sys} restores performance with an average gain of $\mathbf{59.3\%}$, notably achieving up to $89.1\%$ absolute improvement in long-horizon tasks. 
Similarly, on LIBERO-PRO\cite{liberopro}, {\sys} effectively bridges the distribution gap caused by layout and semantic shifts (${\approx}0\%$ success), yielding a weighted average gain of $\mathbf{54.5\%}$. These results confirm {\sys} as a robust, plug-and-play solution for OOD robotics challenges.

% \begin{figure}[htbp]
%      \centering
%      \begin{subfigure}[b]{0.45\textwidth}
%          \centering
%          \includegraphics[width=\textwidth]{figure/remove_distract.pdf}
%          \caption{\textbf{Clutter Removal.} Guided by retrieved experiences, {\sys} masks visual distractors and refines spatial instructions (e.g., from ``far left'' to ``on the left'') to isolate the target for robust grasping.}
%          \label{fig:remove_distract}
%      \end{subfigure}
%      \hspace{1pt}
%      \begin{subfigure}[b]{0.45\textwidth}
%          \centering
%          \includegraphics[width=\textwidth]{figure/prompt_understanding.pdf}
%          \caption{\textbf{Noisy Prompt.} {\sys} rewrites instructions that deviate from retrieved successful-experience patterns into concise commands, improving target identification.}
%          \label{fig:prompt_understanding}
%      \end{subfigure}
%      \hfill
%      \begin{subfigure}[b]{0.45\textwidth}
%          \centering
%          \includegraphics[width=\textwidth]{figure/visual_shift.pdf}
%          \caption{\textbf{Visual Focus.} Leveraging retrieved experiences, {\sys} mitigates attention drift via dual-modality interventions: re-rendering target attributes and simplifying instructions for precise referential grounding.}
%          \label{fig:visual_shift}
%      \end{subfigure}
%      \caption{\textbf{{\sys}-Augmented Manipulation: Case Studies across OOD Challenges.}}
%      \label{fig:effect show}
% \end{figure}

In summary, this work makes three contributions:
\begin{enumerate}
    \item \textbf{Attribution-Driven Memory and Extensible Intervention Architecture:}
    We introduce a closed-loop architecture that overcomes the ``success bias'' of conventional RAG.\@ {\sys} uses contrastive dual-memory retrieval to diagnose failures and dynamically orchestrated MCP tools to execute targeted interventions, forming a parameter-free, plug-and-play robustification framework for VLA models.
    
    \item \textbf{Asynchronous System Design:}
    We decouple task-level intervention from long-term knowledge refinement: the online pipeline executes attribution-guided MCP interventions, while offline consolidation updates attributions and memories to distill reliable priors without interrupting inference or updating parameters.
    
    \item \textbf{In-Context Adaptation and Experimental Results:}
    Across diverse tasks and backbones, {\sys} improves performance and in-context adaptation by shifting reasoning from the frozen policy to an evolving memory base and extensible tools, enabling robustness under OOD conditions where conventional VLA policies typically fail.
    
\end{enumerate}

\vspace{-1em}
\section{Related Work}%
\label{sec:related_work}

\subsection{VLA Models and Generalization Challenges}

Recent years have witnessed a paradigm shift in robot learning, from specialized policy networks to general-purpose VLA models.
Pioneering works such as $\pi_0$~\cite{pi0} and OpenVLA~\cite{openvla} leverage massive web-scale datasets 
(e.g., Open-X Embodiment~\cite{openx}) to enable end-to-end robotic control, 
integrating visual encoders (e.g., SigLIP~\cite{siglip}) with large language models 
to translate open-vocabulary instructions into precise continuous actions.
However, adapting these models to OOD scenarios via fine-tuning 
(e.g., X-VLA~\cite{xvla}) incurs high computational costs, 
motivating the need for parameter-efficient, plug-and-play generalization without weight updates.

\subsection{RAG and In-Context Learning in Robotics}

To improve adaptability without fine-tuning, RAG
and In-Context Learning (ICL) have been introduced to the robotics domain.
Frameworks like RICL~\cite{ricl} and MemER~\cite{memer} retrieve relevant historical demonstrations 
via vector databases and condition policy execution by appending them as contextual prompts, 
enabling adaptation to novel tasks through nearest-neighbor imitation.
While effective at leveraging external memory, these methods focus only on successful trajectories and overlook the diagnostic value of failures, highlighting the need for a dual-memory structure that leverages both successes and failures for contrastive guidance.

\subsection{In-Context Adaptive Agentic Frameworks}

Beyond passive trajectory retrieval, recent research has explored leveraging LLMs and 
VLMs as high-level planners with external tool access.
Code-as-Policies~\cite{cap} generates executable Python code for hierarchical tasks, 
VoxPoser~\cite{voxposer} grounds instructions into 3D spatial value maps for motion planning, 
and MOKA~\cite{moka} annotates affordances on images via visual prompting.
Despite decomposing complex instructions into in-context adaptive executable primitives, 
these frameworks lack the experiential memory and dynamic correction needed for closed-loop adaptation.

\section{Methodology}

\begin{figure*}[t]
  \centering
\vspace{0.5em}  \includegraphics[width=0.85\linewidth]{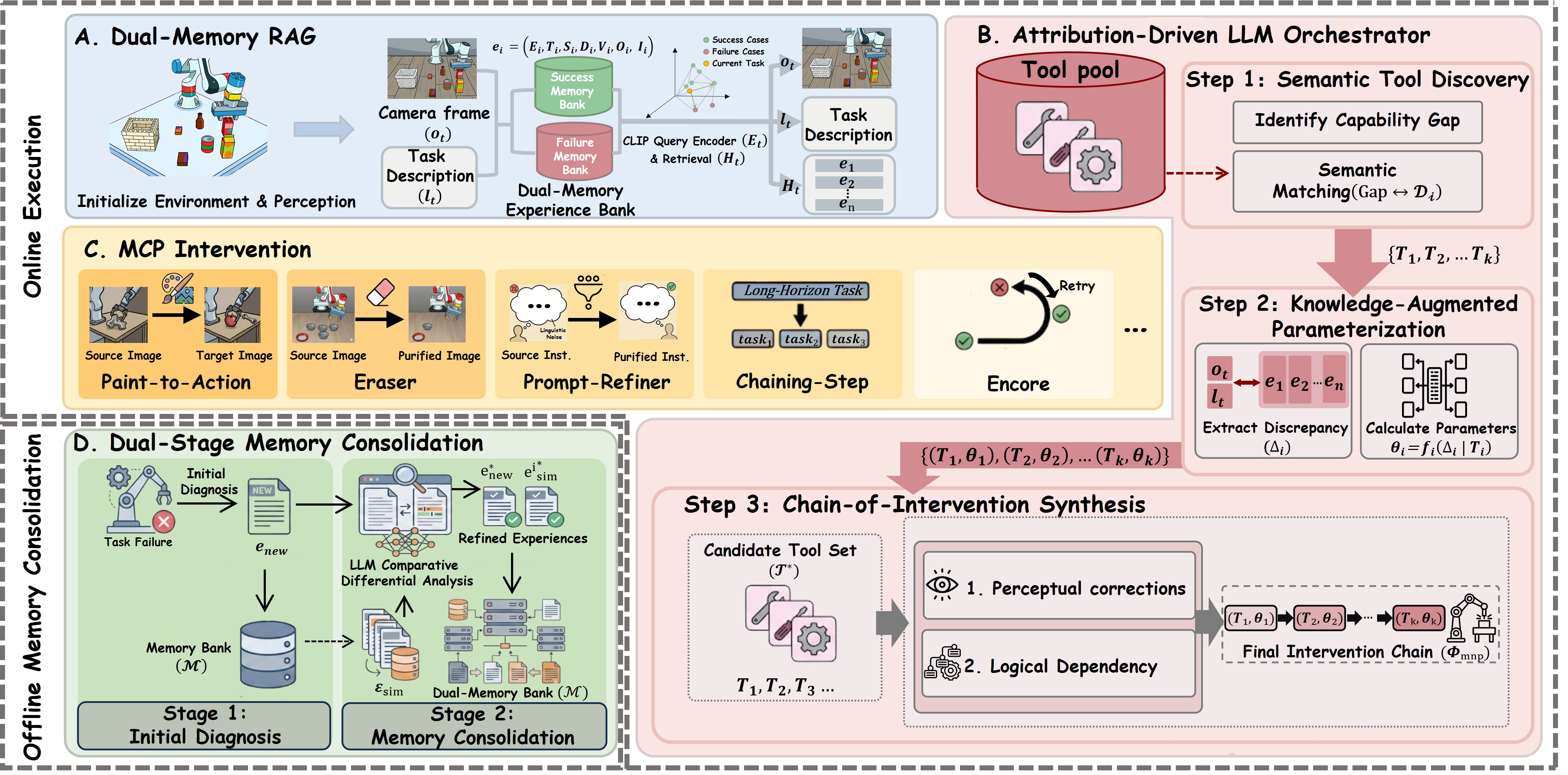}
\caption{\textbf{{\sys} Framework.} Given the observation and instruction, Dual-Memory RAG retrieves relevant experiences from the dual-memory bank. The {\color{black}vision-language orchestrator} selects MCP tools, estimates parameters $\theta$, and synthesizes an intervention chain executed by extensible MCP tools. After execution, asynchronous memory consolidation updates the bank, closing the loop for iterative refinement.}%
\vspace{-1.5em}
  \label{framework}
  
\end{figure*}

This section presents the \textbf{{\sys}} framework, which, as shown in Fig.~\ref{framework}, comprises an \textbf{Online Execution Workflow} and an \textbf{Offline Memory Consolidation Workflow}.

\paragraph{Online Execution Workflow}
The online workflow enables robust task-level adaptation through three modules:
\textbf{(1) Dual-Memory RAG} retrieves success guidance and failure warnings from the Dual-Memory Bank;
\textbf{(2) {\color{black}Attribution-Driven Vision-Language Orchestrator}} diagnoses obstacles and selects attribution-guided interventions;
and \textbf{(3) MCP Intervention} executes them through extensible tools that modulate perceptual and linguistic inputs.

\paragraph{Offline Memory Consolidation Workflow}
Complementing online execution, the offline workflow performs \textbf{Dual-Stage Memory Consolidation}.
It distills newly observed success and failure trajectories into structured memories and updates the Dual-Memory Bank,
enabling continual improvement without parameter updates.

% Task-level components run once per task instance, from initial observation to orchestration, to establish context; 

Together, these workflows form a closed-loop adaptive system.
\vspace{-0.5em}
\[\small
\underbrace{
    \overbrace{\text{Obs.} \rightarrow \text{Retr.} \rightarrow \text{Dec.} \rightarrow \text{Orch.}}^{\substack{\text{Task-level} \\ (\text{once per task})}} \rightarrow \overset{\substack{\text{Chunk-level} \\ (\text{every } N \text{ steps})}}{\text{Exec.}}
}_{\text{Online}} 
\rightarrow 
\underbrace{\text{Consol.}}_{\text{Offline}}
\]

We next describe each module in detail.

\subsection{Dual-Memory RAG}%
\label{dual memory}

Part A of Fig.~\ref{framework} shows a retrieval-augmented Dual-Memory Bank that grounds the {\color{black}orchestrator's reasoning} in prior successes and failures.
Its \textbf{dual} design separates success/failure experiences for contrastive attribution and refines each entry through two-stage consolidation (Section~\ref{Evolution}).

We next detail the bank construction process and the retrieval mechanism built upon it.

\subsubsection{Dual-Memory Storage Design}

We construct a dual-partition memory bank:
\begin{equation}
\mathcal{M} = \mathcal{M}_{succ} \cup \mathcal{M}_{fail},
\end{equation}
where $\mathcal{M}_{succ}$ stores successful interventions as positive guidance, and $\mathcal{M}_{fail}$ stores failed attempts with attribution signals as negative evidence, supporting contrastive reasoning over strategies to reuse and patterns to avoid.

The ablation in Section~\ref{dual memory} shows that single-partition memory yields unstable or inefficient planning, whereas using both partitions reduces reasoning turns and variance.

Each experience entry $e_i \in \mathcal{M}$ is stored in a semi-schema format, consisting of a fixed core schema augmented with extensible auxiliary fields:
\vspace{-0.5em}
\begin{equation}
e_i = \big(
\overbrace{E_i, T_i, S_i, D_i, V_i, O_i}^{\text{fixed schema}}, 
\mathcal{I}_i
\big),
\end{equation}
\begin{itemize}
    \item $E_i$: {\color{black}multimodal} embedding of the task instance;
    \item $T_i$: natural language task description;
    \item $S_i \in \{0,1\}$: success indicator;
    \item $D_i$: diagnostic annotation (non-empty for failures);
    \item $V_i$: keyframe representations of the execution video;
    \item $O_i$: structured scene as object-attribute pairs;
    \item $\mathcal{I}_i$: extensible structured metadata bundle.
\end{itemize}

The metadata bundle $\mathcal{I}_i$ stores tool-execution parameters such as \texttt{key\_frame\_range}, \texttt{success\_max\_step}, and \texttt{rollback}, while allowing new fields as tools or reasoning signals evolve.

\subsubsection{Similarity-Based Retrieval Strategy}

We employ a pre-trained CLIP encoder~\cite{CLIP} $E(\cdot, \cdot)$ to project the current task context (initial observation $o_t$ and instruction $l_t$) into a joint embedding space.
The retrieval is formalized as:
\begin{equation}
\mathcal{H}_t = \text{Top-}k\Big(\text{Sim}\big(E(o_t, l_t), \{E_i \mid e_i \in \mathcal{M}\}\big)\Big),
\end{equation}
$\mathcal{H}_t$ includes success exemplars and failure precedents, providing \textbf{positive guidance} and \textbf{negative counterfactuals} for attribution-driven reasoning.

The tripartite input $(\mathcal{H}_t, o_t, l_t)$ provides the necessary grounding for the {\color{black}orchestrator} to diagnose discrepancies and sequence the appropriate intervention tools.
\vspace{-0.5em}
\subsection{\texorpdfstring{{\color{black}Attribution-Driven Vision-Language Orchestrator}}{Attribution-Driven Vision-Language Orchestrator}}

{\color{black}As shown in Fig.~\ref{framework}B, the Attribution-Driven Vision-Language Orchestrator serves as the deliberative core of {\sys}. It is powered by Qwen3-VL-32B, a multimodal large language model (MLLM) used here as a vision-language model for joint reasoning over observations, instructions, and retrieved visual-semantic memories to diagnose execution failures and orchestrate MCP interventions.}

\subsubsection{Semantic Tool Discovery and Intent Alignment}

The {\color{black}orchestrator} identifies interventions by measuring the capability gap between the current state $o_t$ and retrieved prototypes $\mathcal{H}_t$.
Given semantic tool descriptions $\{\mathcal{D}_i\}_{i=1}^N$, it performs zero-shot matching to select a candidate toolset:
\begin{equation}
    \mathcal{T}_{1:k}= \text{Match}\big(\Delta_{g}, \{\mathcal{D}_i\}_{i=1}^N\big), \quad \Delta_{g} = \text{Gap}(o_t, \mathcal{H}_t),
\end{equation}
selecting tools whose functions align with the diagnosed failure cause.

\subsubsection{Knowledge-Augmented Parameterization}%
\label{Knowledge-Augmented Parameterization}
Following tool selection, the {\color{black}orchestrator} derives execution parameters $\theta_i$ from the {\color{black}multimodal} context discrepancy $\Delta$:
\begin{equation}
 \theta_i = f(\Delta_i \mid T_i), \quad \Delta_i = \text{Gap}(o_t, l_t, \mathcal{H}_t\mid T_i),
\end{equation}
where $f$ denotes the mapping function for tool-specific values. Specifically, $\Delta$ characterizes the deviation of current observation $o_t$ and instruction $l_t$ from historical priors $\mathcal{H}_t$ across visual, semantic, and temporal dimensions. Table~\ref{tab:parameterization_formal} summarizes the mapping from these quantitative discrepancies to the specific execution parameters $\theta_i$ for each tool.
\begin{table}[htbp]
\vspace{-1.5em}
\centering
\caption{Tool-specific Parameterization Mapping.}%
\label{tab:parameterization_formal}
\footnotesize
\renewcommand{\arraystretch}{1.25}
\begin{tabularx}{\columnwidth}{@{} l X l @{}}
\toprule
\textbf{Tool} & \textbf{Discrepancy $\Delta$ from $\mathcal{H}_t$} & \textbf{$\theta_i$} \\
\midrule
Paint  & Feature distance from historical mean.         & $C,\,M,\,A_{mask}$ \\
Eraser & Novel geometry vs.\ background model.          & $D,\,D_{masks}$ \\
Prompt & Instruction mismatch with successful patterns  & $\mathcal{S}_{origin}$ \\
Chain  & Sub-event-level alignment.                     & $(N_{e},N_{b})_k$ \\
Encore & Similarity to historical failure trajectories. & $[s_s,s_e],\,N_w$ \\
\bottomrule
\end{tabularx}
\vspace{3pt}
\begin{flushleft}
\scriptsize
\textbf{$C$}: color/texture;\enspace
\textbf{$M$}: material;\enspace
\textbf{$A_{mask}$}: target-object-mask;\enspace
\textbf{$D$}: distractor descriptions;\enspace
\textbf{$D_{masks}$}: distractor-masks;\enspace
\textbf{$\mathcal{S}_{origin}$}: simplified goal of origin task;\enspace
\textbf{$N_{e,b}$}: execution/buffer steps;\enspace
\textbf{$s_{s,e}$}: start/end steps;\enspace
\textbf{$N_w$}: wait duration.
\end{flushleft}
\vspace{-0.5em}
\end{table}

The final output is a set of parameterized interventions:
\begin{equation}
\mathcal{T}^* = \{(T_1, \theta_1), (T_2, \theta_2), \dots, (T_k, \theta_k)\},
\end{equation}

\subsubsection{Chain-of-Intervention Synthesis}

The {\color{black}orchestrator} then synthesizes the tools into an ordered intervention chain:
\begin{equation}
\begin{aligned}
\Phi_{\text{mcp}} &= \text{Orchestrate}(\mathcal{T}^*) \\
&= \big\langle (T_1, \theta_1) \rightarrow (T_2, \theta_2) \rightarrow \dots \rightarrow (T_k, \theta_k) \big\rangle,
\end{aligned}
\end{equation}

The orchestration follows two principles:
\begin{enumerate}
    \item \textbf{Sensory-First}: Perceptual corrections (e.g., \textit{Paint-to-Action}, \textit{Eraser}) precede physical recovery (e.g., \textit{Encore}) to avoid repeated failures from misperception.
    \item \textbf{Logical Dependency}: Causal-interference removal is applied before downstream reasoning and execution.
\end{enumerate}

\noindent\textbf{For example}, under joint visual shift and execution stagnation, the {\color{black}orchestrator} generates
\textbf{$\text{Paint-to-Action} \rightarrow \text{Encore}$,}
so perceptual rectification precedes rollback. 

\subsection{MCP Intervention}

% \begin{figure}[htbp]
%      \centering
%      \begin{subfigure}[b]{0.45\textwidth}
%          \centering
%          \includegraphics[width=\textwidth]{figure/remove_distract.pdf}
%          \caption{Clutter Removal}
%          \label{fig:remove_distract}
%      \end{subfigure}
%      \hspace{1pt}
%      \begin{subfigure}[b]{0.45\textwidth}
%          \centering
%          \includegraphics[width=\textwidth]{figure/prompt_understanding.pdf}
%          \caption{Noisy Prompt}
%          \label{fig:prompt_understanding}
%      \end{subfigure}
%      \hfill
%      \begin{subfigure}[b]{0.45\textwidth}
%          \centering
%          \includegraphics[width=\textwidth]{figure/visual_shift.pdf}
%          \caption{Visual Focus}
%          \label{fig:visual_shift}
%      \end{subfigure}
%      \caption{{\sys}-Augmented Manipulation: Case Studies}
%      \label{fig:effect show}
% \end{figure}
As depicted in Fig.~\ref{framework}C, this module executes the intervention chain $\Phi_{\text{mcp}}$ through perceptual or physical actions. We instantiate five MCP tools for common VLA failures (e.g., distribution shift and causal distraction) while preserving plug-and-play extensibility for new failure types.

\subsubsection{Paint-to-Action: Visual Domain Shift}
\begin{figure}[htbp]
    \centering
    \vspace{-0.5em}\includegraphics[width=0.95\columnwidth]{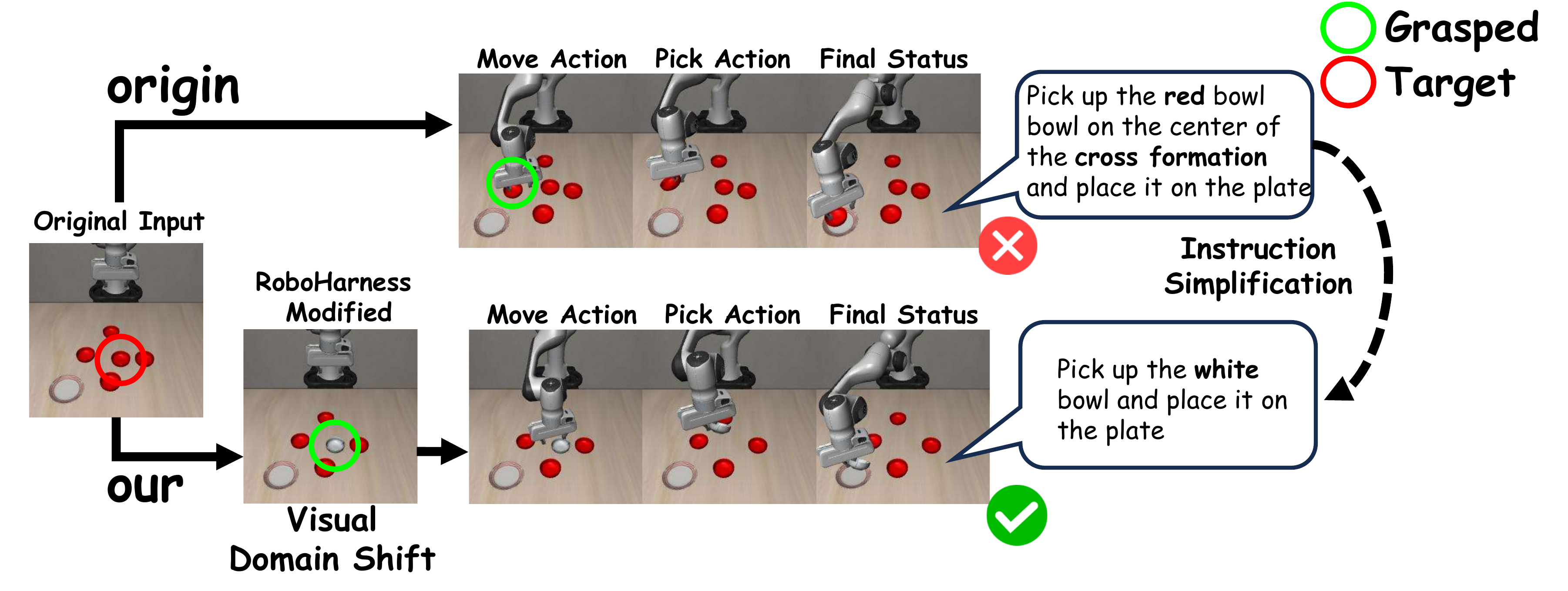}
    \caption{\textbf{Visual Focus.} Addressing visual shift in complex manipulation environments.}%
    \label{fig:visual_shift}
    \vspace{-1em}
\end{figure}
Paint-to-Action performs non-parametric domain adaptation:
\begin{equation}
o_t' = \text{Paint-to-Action}(o_t, \theta_{\text{dist}}),
\end{equation}
Implementation is achieved through \texttt{visual\_overlay} and \texttt{replace\_texture}. The {\color{black}orchestrator} derives $\theta_{\text{dist}}$, SAM3~\cite{sam3} segments target regions, and shift is mitigated by high-contrast masking or textures from successful trajectories (\texttt{rag\_success}) (Fig.~\ref{fig:visual_shift}).

\subsubsection{Eraser: Causal Confusion}
\begin{figure}[htbp]
    \centering
    \vspace{-0.5em}\includegraphics[width=0.95\columnwidth]{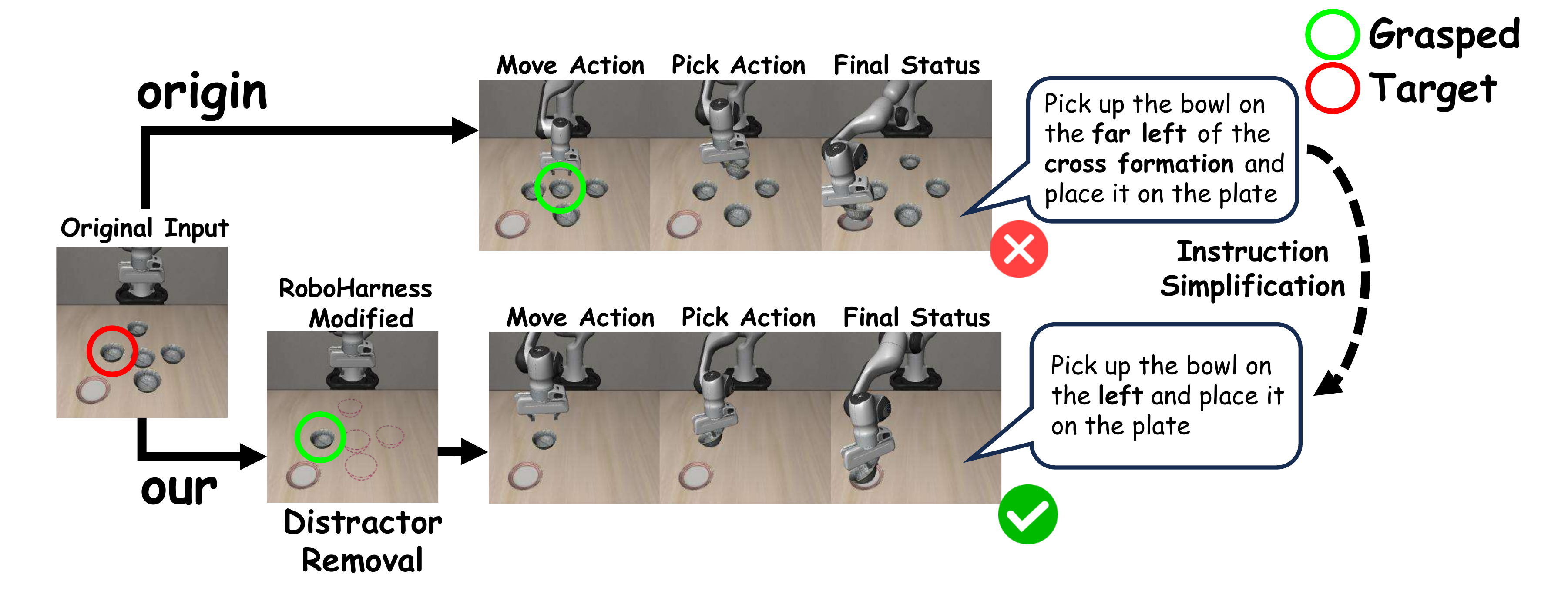}
    \caption{\textbf{Clutter Removal.} Erasing irrelevant distractors to mitigate causal confusion.}%
    \label{fig:remove_distract}
    \vspace{-1em}
\end{figure}
Eraser removes irrelevant objects via $\theta_{\text{conf}}$:
\begin{equation}
o_t' = \text{Eraser}(o_t, \theta_{\text{conf}}),
\end{equation}
Using \texttt{remove\_distractor}, the {\color{black}orchestrator} identifies distractors from $o_t$ and $\mathcal{H}_t$, SAM3~\cite{sam3} extracts masks, and OpenCV Telea inpainting removes them before policy inference (Fig.~\ref{fig:remove_distract}).

\subsubsection{\texorpdfstring{\mbox{Prompt-Refiner: {\color{black}Instruction-Pattern Mismatch}}}{Prompt-Refiner: Instruction-Pattern Mismatch}}
\begin{figure}[htbp]
    \centering
    \vspace{-0.5em}
    \includegraphics[width=0.95\columnwidth]{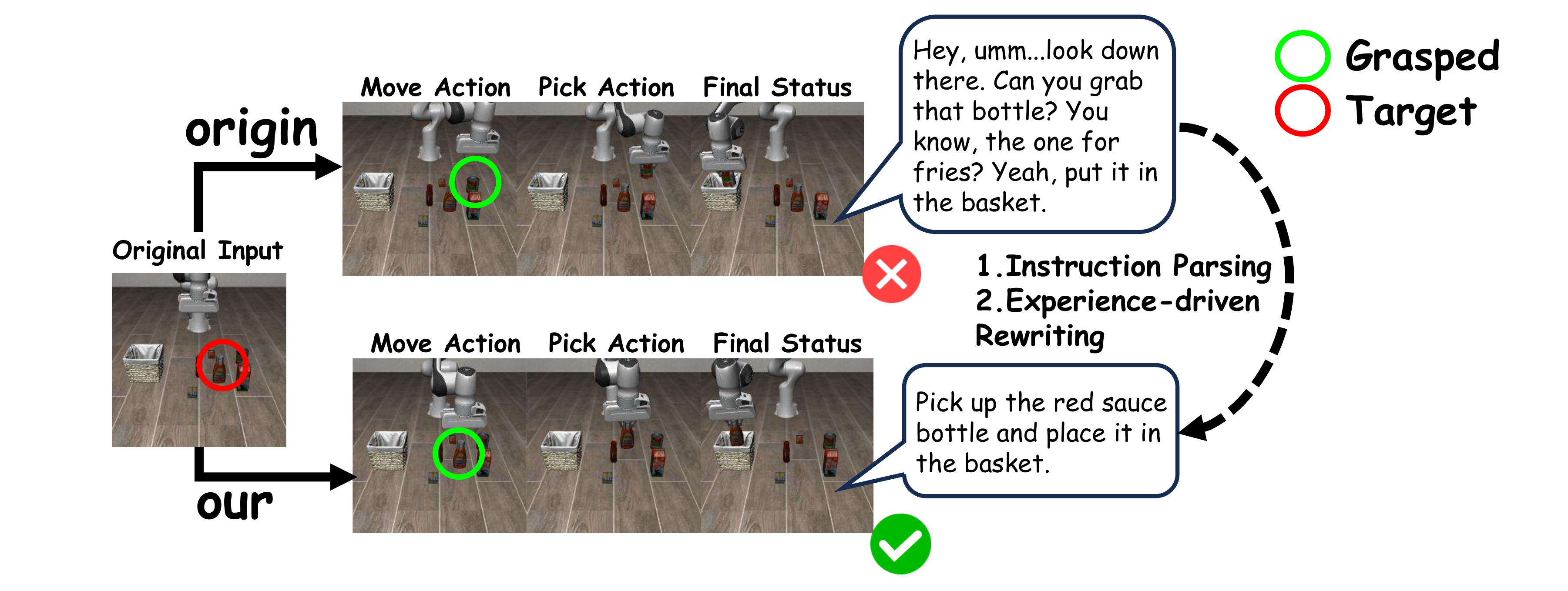}
    \caption{\textbf{Noisy Prompt.} {\color{black}Rewriting instructions that deviate from historical successful experience patterns into concise executable commands.}}%
    \label{fig:prompt_understanding}
    \vspace{-1em}
\end{figure}
Prompt-Refiner {\color{black}refines instructions whose wording or task structure diverges from retrieved successful experiences}:
\begin{equation}
l_t' = \text{Prompt-Refiner}(l_t, \theta_{\text{lang}}),
\end{equation}
where the {\color{black}orchestrator rewrites $l_t$ when its wording or task structure deviates from retrieved successful experience patterns in $\mathcal{H}_t$, producing a concise \texttt{refined\_task} for execution (e.g., colloquial or dialogue-like instructions are rewritten into ``Pick\ldots and place\ldots'' forms in Fig.~\ref{fig:prompt_understanding}).}

\subsubsection{Chaining-Step: Temporal Compounding}

Chaining-Step reduces horizon length:
\begin{equation}
A_{seq}' = \text{Chaining-Step}(A_{seq}, \theta_{\text{chain}}),
\end{equation}
The {\color{black}orchestrator} decomposes macro sequence $A_{seq}$ into aligned sub-tasks using $\mathcal{H}_t$, and assigns temporal bounds (notably $N_e, N_b$) from $\theta_{\text{chain}}$. Replacing $A_{seq}$ with $A_{seq}'$ shortens execution segments and mitigates long-horizon context degradation (e.g., long tasks are split into shorter sub-tasks with ``Reset \& Buffer'' between them in Fig.~\ref{fig:long_horizon}).

\subsubsection{Encore: Execution Stagnation}
\begin{figure}[htbp]
    \vspace{0.5em}
    \centering
    \includegraphics[width=0.95\columnwidth]{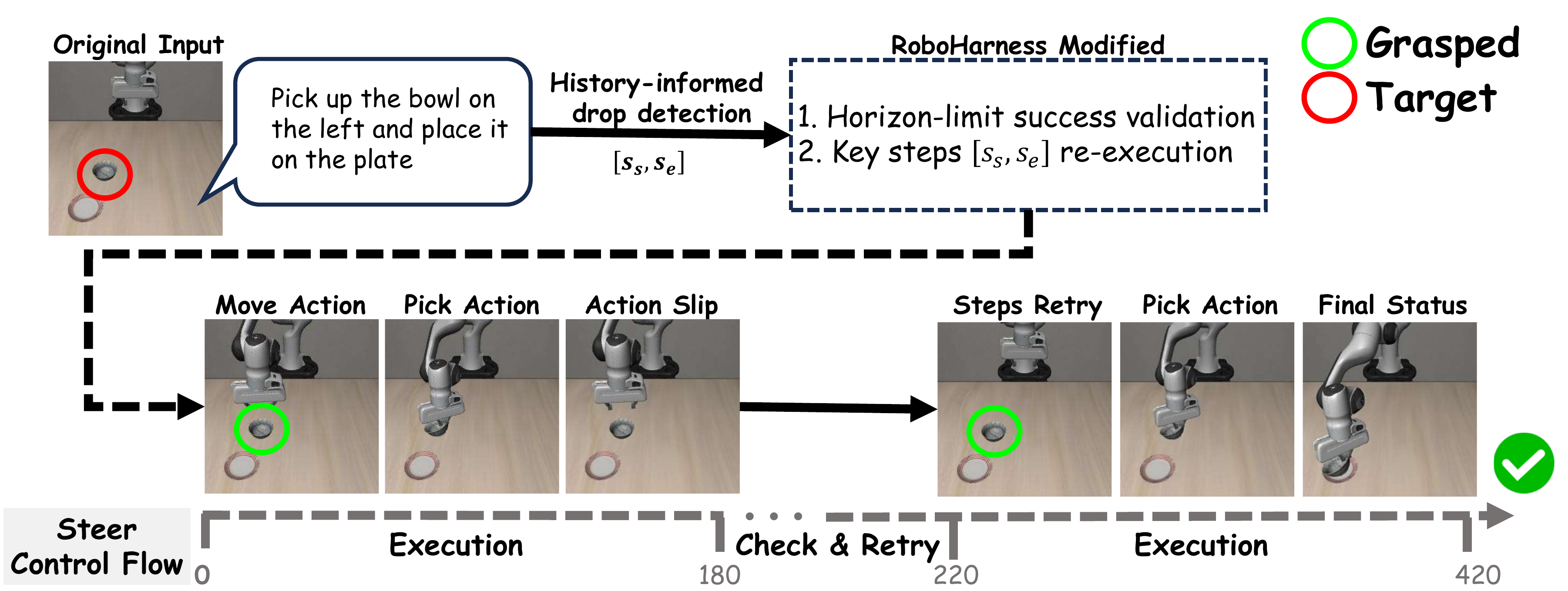}
    \caption{\textbf{Encore.} Recovering via key-step re-execution.}%
    \label{fig:encore}
    \vspace{-1.5em}
\end{figure}
Encore resolves deadlocks via key-state recovery:
\begin{equation}
s'_t = \text{Encore}(s_t, \theta_{\text{stuc}}),
\end{equation}
where $\theta_{\text{stuc}}$ (from failure precedents in $\mathcal{H}_t$) specifies retry bounds $[s_s, s_e]$ and wait time $N_w$. The \texttt{encore} tool resets counters and rolls back via averaged reverse trajectory; if horizon-limit validation fails, it triggers ``Reset \& Retry'' from keyframe $s_s$ (Fig.~\ref{fig:encore}).
\subsection{Dual-Stage Memory Consolidation}%
\label{Evolution}

This module, illustrated in Fig.~\ref{framework}D, consolidates the latest execution trajectory into the Dual-Memory Bank $\mathcal{M}$ using a two-stage attribution mechanism.

\subsubsection{Stage 1: Initial Diagnosis}

Upon task termination, the {\color{black}orchestrator} performs Initial Diagnosis:
\begin{equation}
e_{new} = \text{InitDiag}(o_t, T_{raw}),
\end{equation}
and stores the new experience:
\begin{equation}
\mathcal{M} = \mathcal{M} \cup \{e_{new}\},
\end{equation}

\subsubsection{Stage 2: Memory Consolidation}

Memory Consolidation retrieves the top-$N$ similar experiences $\mathcal{E}_{\text{sim}} = \{e_{\text{sim}}^1, \dots, e_{\text{sim}}^N\}$ for cross-task refinement:
\begin{equation}
\{e_{\text{new}}^*, e_{\text{sim}}^{1*}, \dots, e_{\text{sim}}^{N*}\} = \text{MemConsol}(e_{\text{new}}, \mathcal{E}_{\text{sim}}),
\end{equation}
where each $e^*$ is an updated experience tuple.
The {\color{black}orchestrator} performs batch-level differential analysis to correct historical failure descriptions and optimize correction plans using new evidence, enabling iterative self-correction of $\mathcal{M}$ rather than monotonic expansion.

\vspace{-0.5em}
\section{Experiment}
% \begin{figure*}[t]
%   \centering
%   \includegraphics[width=0.95\linewidth]{figure/result.png}
%     \caption{\textbf{Performance comparison of {\sys} across diverse robotic foundation models and OOD challenges.} The results demonstrate that {\sys} consistently yields substantial performance gains for all base models across visual, linguistic, and long-horizon tasks.}%
%   \label{fig:result}
% \end{figure*}

\begin{figure*}[t]
  \centering
  \vspace{0.5em}
  \begin{minipage}[t]{0.66\textwidth}
    \vspace{0pt}
    \centering
    \includegraphics[width=\linewidth]{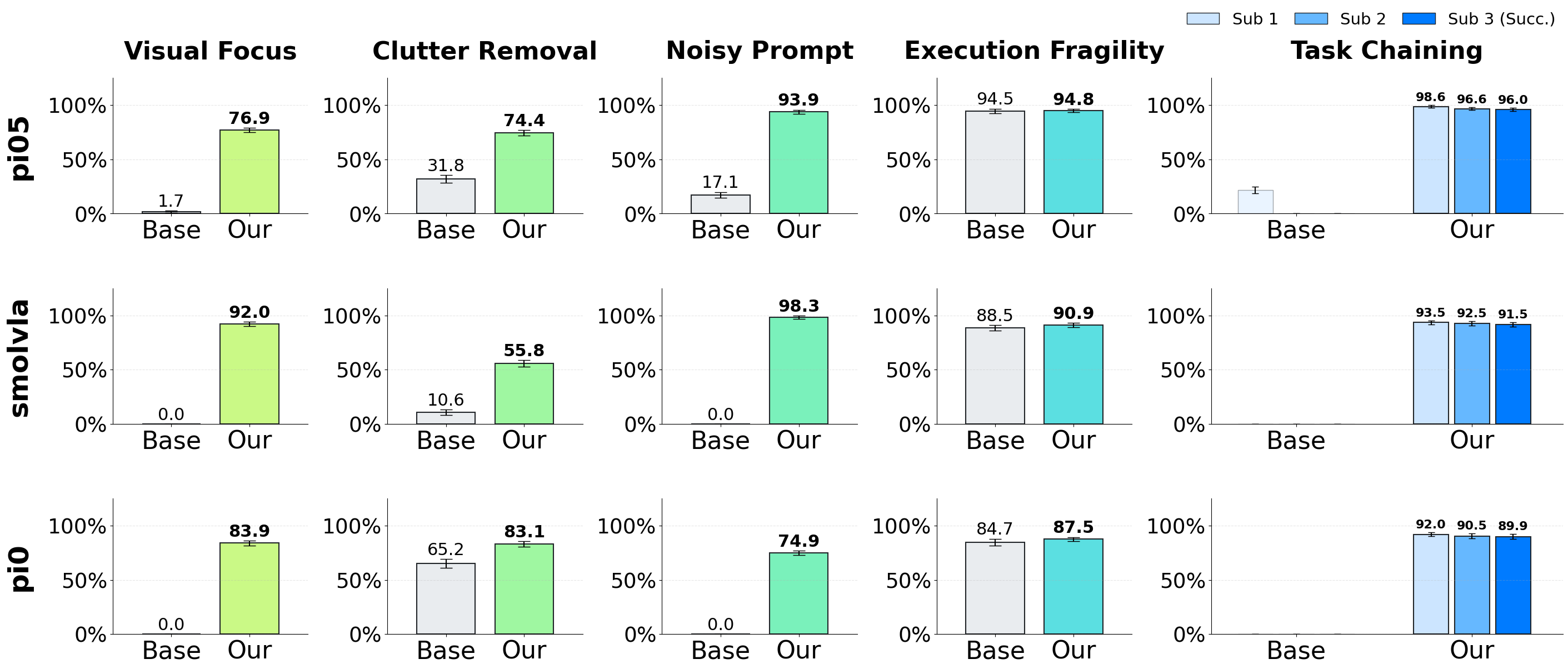}
    \captionof{figure}{\textbf{Performance on LIBERO-{\sys}.} Comparison of {\sys} and VLA baselines across visual, linguistic, and long-horizon challenges.}%
    \label{fig:result}
  \end{minipage}%
  \hspace{0.01\textwidth}%
  \begin{minipage}[t]{0.30\textwidth}
    \vspace{0pt}
    \centering
    {\color{black}
    \vspace{0pt}
    \small
    \setlength{\tabcolsep}{3pt}
    \renewcommand{\arraystretch}{1.02}
    \begin{tabular}{@{}p{0.48\linewidth}lr@{}}
    \toprule
    \multicolumn{1}{@{}l}{\textbf{Component}} & & \textbf{ms/step} \\
    \midrule
    \multirow{2}{=}{Baseline rollout} & Mean & 15.31 \\
    & P95 & 15.98 \\
    \cmidrule(lr){1-3}
    \multirow{2}{=}{RAG retrieval} & Mean & 0.005 \\
    & P95 & 0.005 \\
    \cmidrule(lr){1-3}
    \multirow{2}{=}{VLM orchestration} & Mean & 7.20 \\
    & P95 & 8.65 \\
    \cmidrule(lr){1-3}
    \multirow{2}{=}{SAM3 + Overlay} & Mean & 0.16 \\
    & P95 & 0.16 \\
    \cmidrule(lr){1-3}
    \multirow{2}{=}{SAM3 + Remove} & Mean & 0.16 \\
    & P95 & 0.16 \\
    \bottomrule
    \end{tabular}
    \vspace{-0.3em}
    \captionsetup{justification=raggedright,singlelinecheck=false}
    \vspace{0.2em}
    \captionof{table}{\textbf{Runtime overhead.} Amortized over 1000 LIBERO steps.}
    \label{tab:runtime_overhead}
    \renewcommand{\arraystretch}{1.0}
    }
  \end{minipage}
  \vspace{-0.5em}
\end{figure*}

\begin{table}[t]
  \centering
\caption{\textbf{Performance on LIBERO-PRO.} Success rates under layout (Pos) and semantic (Task) shifts.}%
  \label{tab:libero_pro}
  \small % 调小字号至 footnotesize 以防止冲出栏宽
  \addtolength{\tabcolsep}{-4pt} % 极限压缩列间距
  
  {
  \setlength{\aboverulesep}{0pt}
  \setlength{\belowrulesep}{0pt}
  
  \begin{tabular}{l|clcc}
    \toprule
    \rule{0pt}{3ex} \textbf{Level} & \textbf{Category} & \textbf{Task (Symbolic Form)} & \textbf{$\pi_{0.5}$BASE} & \textbf{w/Our} \\
    \midrule
    
    % --- Position Level ---
    \rule{0pt}{2.5ex} \multirow{6}{*}{\textbf{Pos}} & \multirow{5}{*}{\textbf{Spat.}} 
      & $P.(btw(p, r), pl.)$        & 2\% & \textbf{63.0\%} \\
    & & $P.(center, pl.)$            & 0\% & \textbf{97.0\%} \\
    & & $P.(drw_{top}(cab_{w}), pl.)$ & 0\% & \textbf{20.2\%} \\
    & & $P.(next\_to(p), pl.)$       & 0\% & \textbf{57.0\%} \\
    & & $P.(next\_to(r), pl.)$       & 12\% & \textbf{71.0\%} \\
    \cmidrule(l){2-5}
    & \rule{0pt}{2.5ex} \textbf{Obj.} & $P.(soup, bskt.)$ & 0\% & \textbf{35.0\%} \\
    \midrule
    % 修复后的行：把 \rule 放在 \multicolumn 内部
    \multicolumn{3}{l|}{\rule{0pt}{2.5ex} \textit{\textbf{Average (Pos)}}} & 2.33\% & \textbf{57.2\%} \\
    
    \midrule 
    
    % --- Task Level ---
    \rule{0pt}{3ex} \multirow{14}{*}{\textbf{Task}} & \multirow{9}{*}{\textbf{Spat.}} 
      & $P.(btw(p, r), pl.)$          & 0\% & \textbf{78.0\%} \\
    & & $P.(center, pl.)$              & 2\% & \textbf{82.0\%} \\
    & & $P.(next\_to(ck.), pl.)$    & 2\% & \textbf{97.0\%} \\
    & & $P.(next\_to(p), pl.)$        & 0\% & \textbf{73.0\%} \\
    & & $P.(next\_to(r), pl.)$        & 2\% & \textbf{98.0\%} \\
    & & $P.(on(ck.), pl.)$          & 0\% & \textbf{12.0\%} \\
    & & $P.(on(r), pl.)$               & 2\% & \textbf{100.0\%} \\
    & & $P.(on(stove), pl.)$           & 0\% & \textbf{13.0\%} \\
    & & $P.(on(cab_{w}), pl.)$        & 0\% & \textbf{95.0\%} \\
    \cmidrule(l){2-5}
    & \rule{0pt}{2.5ex} \multirow{5}{*}{\textbf{Obj.}} 
      & $P.(soup, bskt.)$              & 0\% & \textbf{16.0\%} \\
    & & $P.(bbq, bskt.)$               & 2\% & \textbf{32.0\%} \\
    & & $P.(juice, bskt.)$             & 2\% & \textbf{41.0\%} \\
    & & $P.(dressing, bskt.)$          & 0\% & \textbf{13.0\%} \\
    & & $P.(tomato, bskt.)$            & 0\% & \textbf{23.0\%} \\
    \midrule
    % 同理修复
    \multicolumn{3}{l|}{\rule{0pt}{2.5ex} \textit{\textbf{Average (Task)}}} & 0.86\% & \textbf{55.2\%} \\
    \bottomrule
  \end{tabular}
  }
  \vspace{-1.5em}
\end{table}
We evaluate {\sys} across multiple foundation models on 25 tasks. We further conduct ablation studies to validate the Dual-Stage Memory Consolidation mechanism.
Results show that {\sys} consistently improves in-context adaptation and policy robustness across scenarios.

To improve statistical reliability under real-robot cost and trial constraints, we prioritize rigorous testing in a high-fidelity simulation environment.

\subsection{Experimental Setup}
\noindent\textbf{Task Design and Evaluation Benchmarks.}
To evaluate robust in-context adaptation under distribution shifts, we conduct experiments across two task suites. We first adopt the \textbf{LIBERO-PRO}~\cite{liberopro} benchmark to test robustness against Position-swap (LIBERO-PRO Pos) and Task-level (LIBERO-PRO Task) variations within Spatial and Object categories.

Furthermore, we introduce \textbf{LIBERO-{\sys}}. This suite formalizes five \textit{prevalent and critical} OOD challenge dimensions across simulated and real-world robotics. By reconfiguring assets within LIBERO scenarios~\cite{libero_ref}, it provides a controlled environment to assess an agent's ability to trigger adaptive interventions for recurring failure modes without task-specific fine-tuning.

\begin{itemize}
    \item \textbf{Visual Focus}: Selecting a target from five identical, novel bowls. By substituting \textit{LIBERO-Spatial} targets with indistinguishable assets, we evaluate \textbf{referential grounding} under fine-grained visual ambiguity.

    \item \textbf{Clutter Removal}: Grasping targets amidst dense distractors. We augment \textit{LIBERO-Spatial} with high-entropy object clusters to test \textbf{operational focus} and interference mitigation in crowded scenes.

    \item \textbf{Noisy Prompt}: {\color{black}Intent extraction from instructions whose wording or task structure deviates from historical successful experience patterns} (e.g., \textit{``Umm\ldots get that red squeezy thing''}). By injecting linguistic noise into \textit{LIBERO-Object} instructions, we assess \textbf{linguistic robustness} and {\color{black}experience-grounded semantic mapping}.

    \item \textbf{Execution Fragility}: Critical maneuvers where minor slips trigger failure. We curate unstable sub-tasks from \textit{LIBERO-Object} to evaluate \textbf{operational stability} against mechanical uncertainties.

    \item \textbf{Task Chaining}: Long-horizon sorting of multiple objects into a basket. By sequencing atomic \textit{LIBERO-Object} actions, this task evaluates \textbf{long-term reasoning} and \textbf{error resilience} in sequential workflows.
\end{itemize}

% \noindent\textbf{Complex Spatial and Object Reasoning.}
% Beyond the aforementioned OOD challenges, we further evaluate the model on the \textbf{LIBERO-PRO}~\cite{liberopro} benchmark, testing its robustness against Position-swap (\textbf{Pos}) and Task-level (\textbf{Task}) variations within \textbf{Spatial} and \textbf{Object} tasks. 

\noindent\textbf{In-Context Initialization and Metrics.}
We initialize Dual-Memory RAG with a preliminary inference pass on standard LIBERO-Object and LIBERO-Spatial sets.
This non-parametric ``cold start'' populates the memory bank with generic execution traces, without gradient-based training or exposure to evaluation-time OOD perturbations.
We then run 199 randomized rollouts per task, fully shuffling object poses and task sequences to reduce bias.
Performance is measured by \textbf{overall completion rates} and \textbf{intermediate success rates} for granular long-horizon analysis.

\noindent\textbf{Baselines.}
We evaluate {\sys} on diverse foundation models, including $\pi_0$~\cite{pi0}, $\pi_{0.5}$~\cite{pi05}, and SmolVLA~\cite{SmolVLA}.
Across OOD scenarios, {\sys} consistently improves success rates, indicating better policy generalization.

\subsection{Main Results}%
\label{Main Results}

% \begin{figure*}[t]
%   \centering
%   \includegraphics[width=0.8\linewidth]{figure/result.png}
%     \caption{\textbf{Performance comparison of {\sys} across diverse robotic foundation models and OOD challenges.} The results demonstrate that {\sys} consistently yields substantial performance gains for all base models across visual, linguistic, and long-horizon tasks.}
%   \label{fig:result}
% \end{figure*}
As shown in Fig.~\ref{fig:result} and Table~\ref{tab:libero_pro}, {\sys}-augmented policies improve robustness on both LIBERO-{\sys} and LIBERO-PRO, especially under perceptual ambiguity, linguistic noise, and long-horizon compounding errors.

\subsubsection{Results on LIBERO-{\sys}}

\noindent\textbf{Environmental and Linguistic Variations.}
Base models are sensitive to sensory and linguistic interference.
For \textit{pi05-LIBERO}, \textit{Visual Focus} and \textit{Clutter Removal} success rates increase from $1.7\%$ and $31.8\%$ to $76.9\%$ and $74.4\%$ with {\sys}, showing MCP-based interventions help isolate task-relevant objects.
For language perturbations, \textit{smolvla-LIBERO} reaches $98.3\%$ on \textit{Noisy Prompt} after instruction refinement.

\noindent\textbf{Execution Stability and Long-Horizon Tasks.}
In \textit{Execution Fragility}, {\sys} raises success rates above $87\%$ for all models by adding recovery-oriented interventions.
{\color{black}Since the baseline success rates in \textit{Execution Fragility} are already relatively high, {\sys}'s improvement is smaller in absolute terms and mainly comes from reducing the remaining rare failures through rollback and retry.}
In \textit{Task Chaining}, {\sys} reduces compounding errors through subtask decomposition and control-flow reset; notably, \textit{pi05-LIBERO} maintains $96.0\%$ success through the final sub-task.

\subsubsection{Results on LIBERO-PRO}

\noindent\textbf{Position-Level Layout Shifts.}
Under LIBERO-PRO Pos, the $\pi_{0.5}$ baseline achieves only $2.33\%$ average success, while {\sys} improves it to $57.2\%$.
{\color{black}This suggests that memory-guided visual interventions help recover object grounding under layout shifts.}

\noindent\textbf{Task-Level Semantic Perturbations.}
Under LIBERO-PRO Task, the baseline averages only $0.86\%$ success, whereas {\sys} reaches $55.2\%$.
{\color{black}These gains indicate that {\sys} improves the input and task context for the frozen controller, rather than replacing the low-level VLA policy.}

{\color{black}
\subsubsection{Discussion}
The large gains stem from compensating for frozen VLAs' statelessness under OOD layouts, distractors, noisy prompts, and long-horizon chains.
{\sys} retrieves success/failure memories to diagnose failures and trigger task-level MCP interventions before execution.
}

{\color{black}
\subsubsection{Online Latency}
Table~\ref{tab:runtime_overhead} reports latency amortized over a 1000-step LIBERO task, with baseline rollout measured on an RTX 4090.
RAG retrieval and SAM3-based tools add negligible amortized overhead, while VLM orchestration is the main extra cost.
Because {\sys} invokes these modules before rollout rather than at every policy step, low-level execution remains smooth and non-blocking.
In pipelined multi-task settings, next-task orchestration can overlap with current task execution.
}

\subsection{Ablation Study}

With the {\color{black}orchestrator + MCP modules} validated in Section~\ref{Main Results}, we further analyze the \textbf{Dual} design of our Dual-Memory RAG system. Following Section~\ref{dual memory ablation}, we proceed in two steps: we first isolate the Dual-Memory Bank's effect on execution efficiency, then evaluate the contribution of Dual-Stage Memory Consolidation to reasoning depth.

\subsubsection{Efficacy of the Dual-Memory Bank}%
\begin{figure}[htbp]
   \vspace{0.5em}
    \centering
    \includegraphics[width=0.4\textwidth]{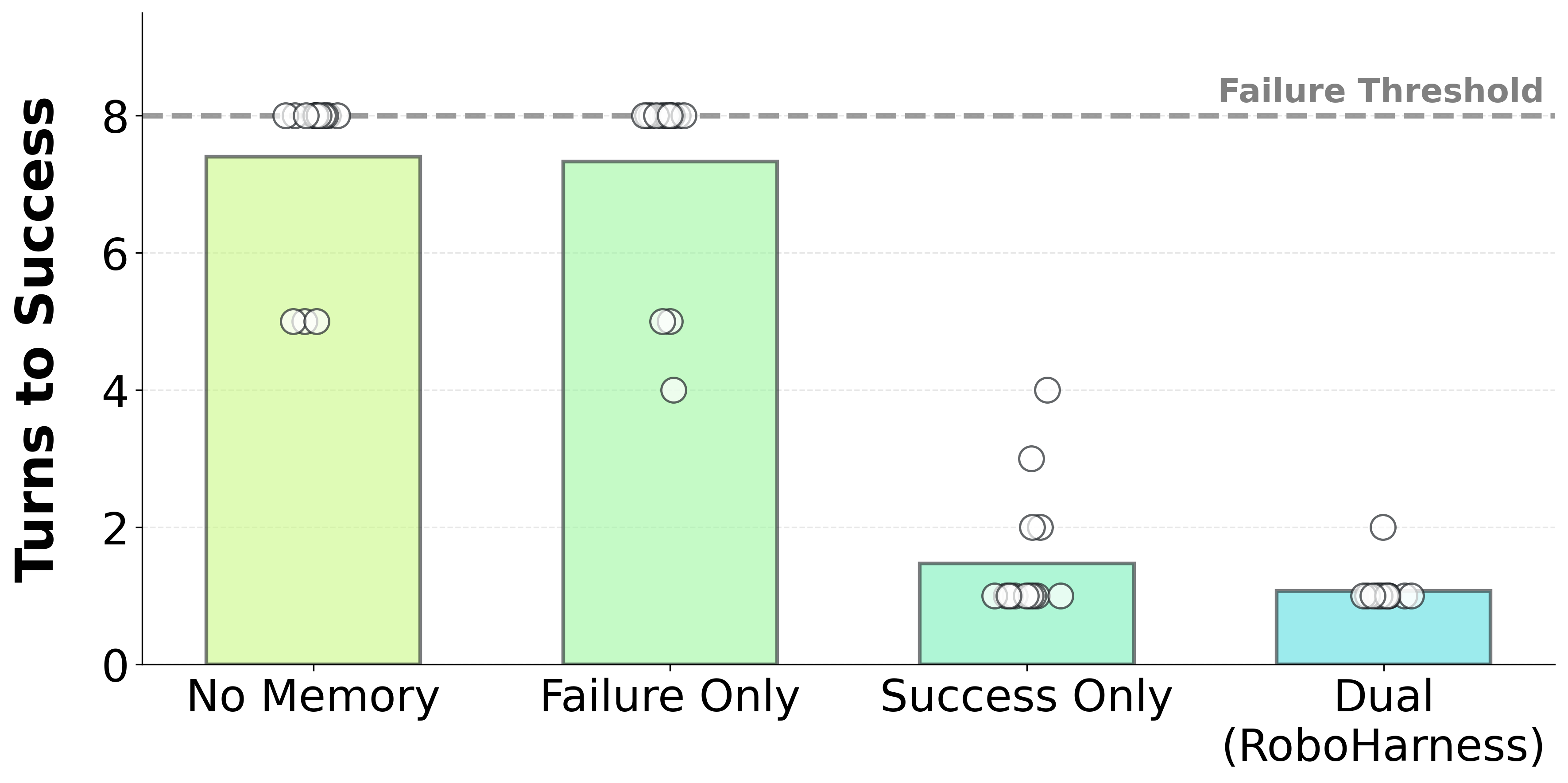}
    \caption{\textbf{Ablation Study.} Distribution of reasoning turns required to reach the target intervention plan.}%
    \label{fig:ablation_beeswarm}
    \vspace{-0.5em}

\end{figure}

\label{dual memory ablation}
To verify the need for both successful and failed experiences in anchoring {\color{black}orchestrator decisions}, we use Average Turns-to-Success~\cite{multiwoz,toolsandbox}, which measures the number of reasoning-intervention loops required to reach an optimal plan.
When an intervention fails, the agent receives sparse environmental feedback and must self-correct in subsequent turns.
As shown in Fig.~\ref{fig:ablation_beeswarm}, we compare four memory configurations: \textbf{No Memory}, \textbf{Failure Only}, \textbf{Success Only}, and \textbf{{\sys} (Dual)}.

As shown in Fig.~\ref{fig:ablation_beeswarm}, the baseline without memory consistently reaches the maximum execution timeout ($Mean = 7.40$), indicating that zero-shot reasoning is insufficient for OOD recovery. While \textit{Success-Only} memory reduces the average turns to $1.47$, it still exhibits significant stochastic variance. In contrast, \textbf{{\sys} (Dual)} converges to a near-optimal \textbf{$1.07$} turns. The dual-memory bank constrains the search space for the {\color{black}vision-language orchestrator}, enabling directed decision-making instead of stochastic exploration.

\subsubsection{Impact of Dual-Stage Memory Consolidation} 
While the dual-memory bank provides the raw data, the quality of these experiences is governed by the iterative attribution module described in Section~\ref{Evolution}. We evaluate this process via three configurations: \textbf{No RAG} (baseline), \textbf{Limited RAG} (single-stage online attribution), and \textbf{Rich RAG} ({\sys}'s dual-stage consolidation).

% --- Table Configuration ---
\newcolumntype{M}[1]{>{\centering\arraybackslash}m{#1}}
\renewcommand{\arraystretch}{1.1}
\setlength{\tabcolsep}{3pt}

\begin{table*}[t]
\vspace{0.5em}
\centering
\caption{\textbf{Ablation Study.} Comparison of reasoning logic and success rates across RAG configurations.}%
\label{ablation}
\footnotesize 
\begin{tabularx}{\textwidth}{|M{1.8cm}|X|M{3.5cm}|M{2.1cm}|M{1.8cm}|M{1.1cm}|}
\hline
\textbf{RAG Config} & \multicolumn{1}{c|}{\textbf{{\color{black}Orchestrator Reasoning via RAG}}} & \textbf{MCP Tool} & \textbf{Action Result} & \textbf{Task Image} & \textbf{SR (\%)} \\ \hline
\multirow{3}{*}{No-RAG} & Ambiguity in target ID due to identical textures. & $o_t' = \text{Paint-to-Action}(o_t, \theta_{\text{dist}})$ & Highlight bowl & \multirow{3}{*}{\includegraphics[width=0.95cm, valign=m]{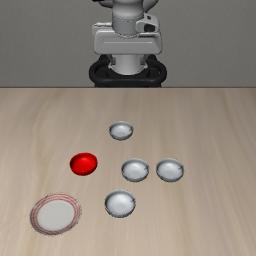}} & \multirow{3}{*}{19.2} \\ \cline{2-4}
 & Potential detection errors from plate shadows. & \texttt{none} & --- & & \\ \cline{2-4}
 & Clean background, no distractors found. & \texttt{none} & --- & & \\ \hline
\multirow{4}{*}{Limited-RAG} & Edge proximity causes occlusion; highlight target. & $o_t' = \text{Paint-to-Action}(o_t, \theta_{\text{dist}})$ & Highlight bowl & \multirow{4}{*}{\includegraphics[width=0.95cm, valign=m]{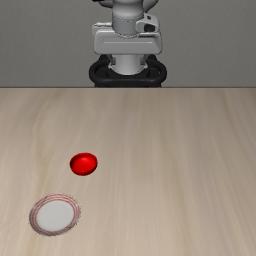}} & \multirow{4}{*}{48.8} \\ \cline{2-4}
 & 1. \textbf{Past failures} suggest high mis-ID risk. & \multirow{2}{*}{$o_t'' = \text{Eraser}(o_t', \theta_{\text{conf}})$} & \multirow{2}{*}{Remove 4 dist.} & & \\ 
 & 2. Focus by \textbf{removing} identical distractors. & & & & \\ \cline{2-4}
 & \textbf{Background replace} to reduce visual noise. & $o_t''' = \text{Eraser}(o_t'', \theta_{\text{conf}})$ & BG Replacement & & \\ \hline
\multirow{5}{*}{Rich-RAG} & 1. High risk from \textbf{distractors via failure memory}. & \multirow{4}{*}{$o_t' = \text{Eraser}(o_t, \theta_{\text{conf}})$} & \multirow{4}{*}{Remove 4 dist.} & \multirow{5}{*}{\includegraphics[width=0.95cm, valign=m]{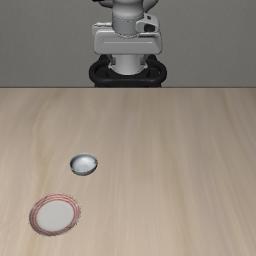}} & \multirow{5}{*}{60.1} \\ 
 & 2. \textbf{Memory} dictates explicit removal of each bowl. & & & & \\ 
 & 3. Strategic preservation of target and plate. & & & & \\ 
 & 4. Use precise position to guide policy. & & & & \\ \cline{2-4}
 & \textbf{Distraction-free; no visual overlay needed.} & \texttt{none} & --- & & \\ \hline
\end{tabularx}

\end{table*}

\begin{figure}[htbp]

     \centering
     \includegraphics[width=0.4\textwidth]{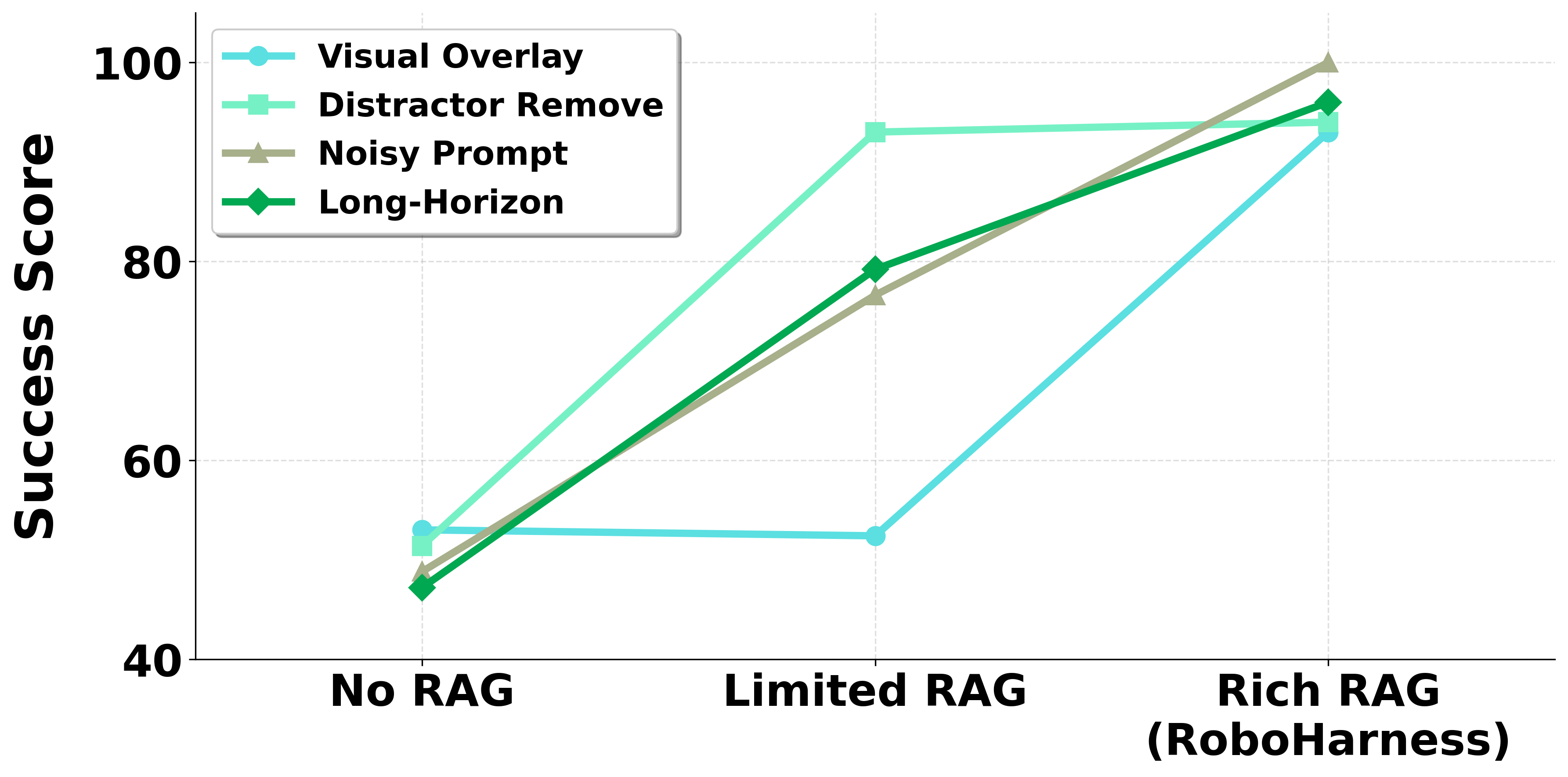}
    \caption{\textbf{Ablation Study.} Evolution of qualitative reasoning depth scores across configurations.}%
    \label{score_qwen} 

\end{figure}

\noindent\textbf{Qualitative Evaluation of Reasoning Depth.}
We use Qwen3-VL-32B to score the logical coherence of generated instructions.
As shown in Fig.~\ref{score_qwen}, Rich RAG consistently achieves the highest ratings across all dimensions.
This gain comes from more focused reasoning trajectories that prioritize critical intervention points and reduce ineffective actions.
These results suggest that higher-quality RAG data improves causal understanding and yields a more robust tool-invocation chain, consistent with the quantitative gains.

\noindent\textbf{Quantitative Performance and Stability.}
We evaluate a ``Clutter Removal'' variant with $\pi_{0.5}$.
As shown in Table~\ref{ablation}, success rates increase with RAG richness.
No RAG achieves only $19.20\%$, and comparison with Section~\ref{Main Results} shows that {\color{black}orchestrator guidance} without RAG can be counterproductive by introducing negative bias.
Limited RAG improves performance to $48.80\%$ but still relies on redundant steps (e.g., unnecessary background replacement).
Rich RAG ({\sys}) reaches $60.10\%$ with a streamlined reasoning chain, showing higher strategic efficiency and precision.

\section{CONCLUSIONS}

\rev{In the tested OOD settings, many VLA failures manifest as perceptual, linguistic, or coordination limitations rather than purely motor deficiencies.}
To address this, we propose {\sys}, \rev{a memory-augmented policy harness}
\rev{that wraps frozen VLA policies via dynamic MCP tool orchestration and a closed-loop cycle of retrieval,}
attribution-driven reasoning, and dual-stage memory consolidation, without any parameter fine-tuning.
Evaluations across multiple VLA backbones show that {\sys} consistently 
improves in-context adaptation and robustness under diverse OOD conditions.
Future work will expand the MCP tool repository to cover a broader range of failure modes, 
investigate frontier multimodal models to further refine failure attribution granularity,
and study how such memory-guided interventions transfer to more diverse real-world robotic platforms.

\section*{ACKNOWLEDGMENT}

We sincerely thank the anonymous reviewers, whose reviews, feedback, and
suggestions have significantly strengthened our work. This
work was supported in part by New Generation Information Technology Program from
Shanghai Committee of Science and Technology (No. 25511104100), National Natural
Science Foundation of China (No. 62432010), and the Fundamental and
Interdisciplinary Disciplines Breakthrough Plan of the Ministry of Education of
China (JYB2025XDXM122). This work was also supported by OpenHarmony Embodied AI
Project Management Committee (PMC).

\bibliographystyle{IEEEtran}
\bibliography{ref}

\end{document}